\def\year{2022}\relax
\documentclass[letterpaper]{article} 
\usepackage{aaai22}  
\usepackage{times}  
\usepackage{helvet}  
\usepackage{courier}  
\usepackage[hyphens]{url}  
\usepackage{graphicx} 
\usepackage{natbib}  
\usepackage{caption} 
\DeclareCaptionStyle{ruled}{labelfont=normalfont,labelsep=colon,strut=off} 
\usepackage{xcolor}

\usepackage{amsmath,amsfonts,bm}

\def\eqref#1{equation~\ref{#1}}

\def\1{\bm{1}}

\DeclareMathAlphabet{\mathsfit}{\encodingdefault}{\sfdefault}{m}{sl}
\SetMathAlphabet{\mathsfit}{bold}{\encodingdefault}{\sfdefault}{bx}{n}

\DeclareMathOperator*{\argmax}{arg\,max}

\usepackage{acronym}
\acrodef{NLU}{natural language understanding}
\acrodef{NLP}{natural language processing}
\acrodef{BNN}{Bayesian neural network}
\acrodef{MC}{Monte Carlo}
\acrodef{RNN}{recurrent neural network}
\acrodef{ELBO}{evidence lower bound}
\acrodef{CR}{customer review}
\acrodef{MCC}{Matthews correlation coefficient}
\usepackage[ruled,linesnumbered,noend]{algorithm2e}

\newcommand{\bfx}{\mathbf{x}}

\newcommand{\bW}{\mathbf{W}}
\newcommand{\bR}{\mathbb{R}}

\usepackage{cancel} 

\newtheorem{proposition}{Proposition}[section]
\newtheorem{proof}{Proof}[section]

\usepackage{newfloat}
\usepackage{listings}
\lstdefinestyle{customc}{
  belowcaptionskip=1\baselineskip,
  breaklines=true,
  frame=L,
  xleftmargin=\parindent,
  language=Python,
  showstringspaces=false,
  basicstyle=\tiny \ttfamily,
  keywordstyle=\bfseries\color{green!40!black},
  commentstyle=\itshape\color{purple!40!black},
  identifierstyle=\color{blue},
  stringstyle=\color{orange},
}
\DeclareCaptionFont{white}{ \color{white} }
\DeclareCaptionFormat{listing}{
  \colorbox[cmyk]{0.43, 0.35, 0.35,0.01 }{
    \parbox{\textwidth}{\hspace{15pt}#1#2#3}
  }
}
\title{Transformer Uncertainty Estimation with Hierarchical Stochastic Attention}
\author {
    Jiahuan Pei\textsuperscript{\rm 1,2}\thanks{This work has been done while doing internship at Amazon.},
    Cheng Wang\textsuperscript{\rm 2}\thanks{Corresponding author.},
    Gy\"{o}rgy Szarvas\textsuperscript{\rm 2}
}
\affiliations {
    \textsuperscript{\rm 1}University of Amsterdam\\
    \textsuperscript{\rm 2}Amazon Development Center Germany GmbH, Berlin, Germany\\
    \{jpei, cwngam, szarvasg\}@amazon.de
}

\usepackage{bibentry}

\begin{document}

\maketitle

\begin{abstract}
Transformers are state-of-the-art in a wide range of NLP tasks and have also been applied to many real-world products.
Understanding the reliability and certainty of transformer model predictions is crucial for building trustable machine learning applications, e.g., medical diagnosis. 
Although many recent transformer extensions have been proposed, the study of the uncertainty estimation of transformer models is under-explored. 
In this work, we propose a novel way to enable transformers to have the capability of uncertainty estimation and, meanwhile, retain the original predictive performance. 
This is achieved by learning a hierarchical stochastic self-attention that attends to values and a set of learnable centroids, respectively.
Then new attention heads are formed with a mixture of sampled centroids using the Gumbel-Softmax trick. 
We theoretically show that the self-attention approximation by sampling from a Gumbel distribution is upper bounded. 
We empirically evaluate our model on two text classification tasks with both in-domain (ID) and out-of-domain (OOD) datasets.
The experimental results demonstrate that our approach: 
(1) achieves the best predictive performance and uncertainty trade-off among compared methods;
(2) exhibits very competitive (in most cases, improved) predictive performance on ID datasets; 
(3) is on par with Monte Carlo dropout and ensemble methods in uncertainty estimation on OOD datasets.
\end{abstract}

\section{INTRODUCTION}
\begin{figure}[!htb]
\centering
\includegraphics[width=0.48\textwidth]{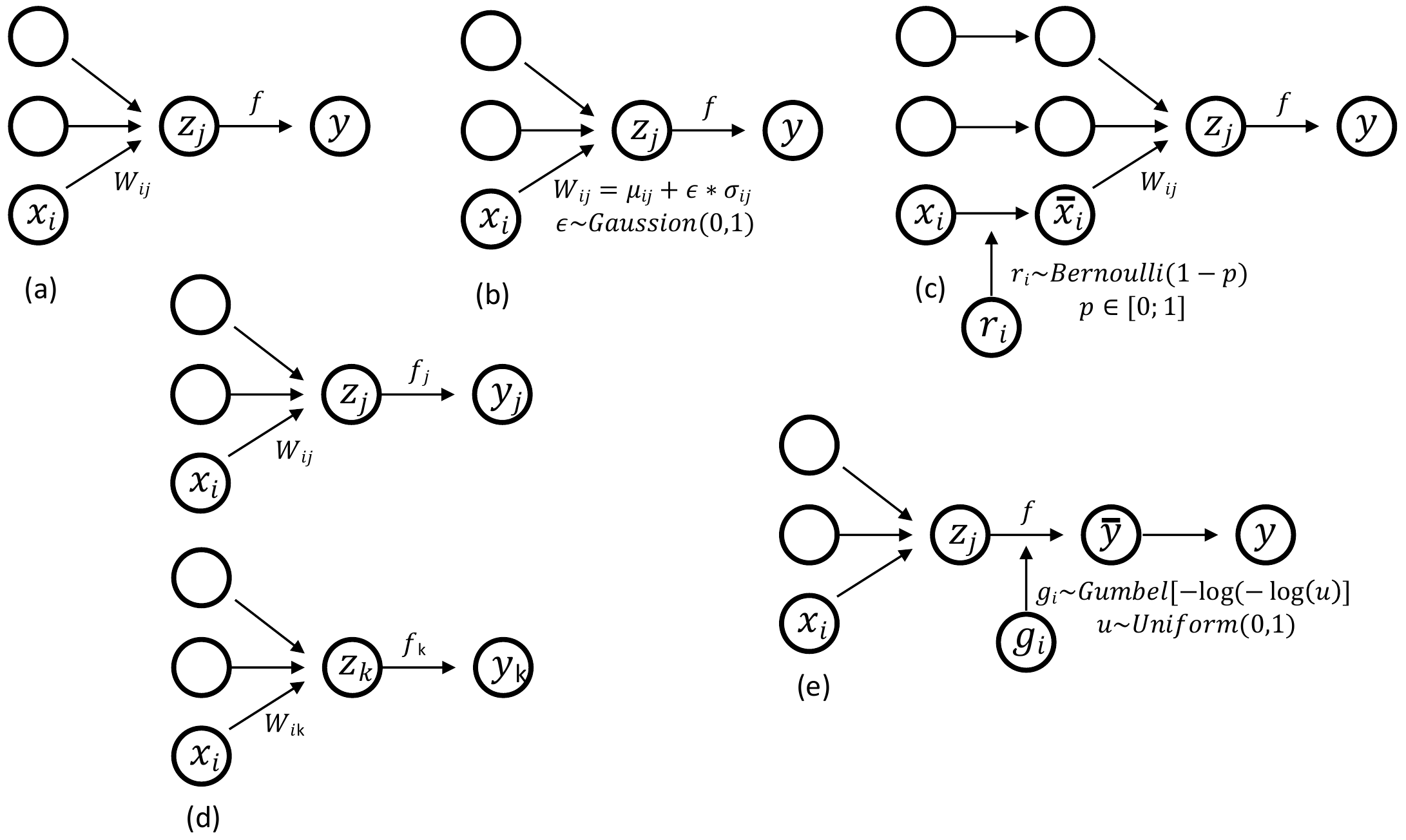}
\caption{The methods of uncertainty estimation. 
(a) Deterministic neural network outputs a single-point prediction; 
(b) Bayesian neural network captures uncertainty via sampling from a Gaussian distribution; 
(c) Variational dropout captures uncertainty via sampling dropout masks from a Bernoulli distribution; 
(d) Ensemble captures uncertainty by combining multiple independently trained deterministic models with different random seeds; 
(e) Gumbel-Softmax trick for uncertainty estimation, the randomness comes from the sampling categorical distribution from a Gumbel.}
\label{fig:uncertainty}
\vspace*{-1\baselineskip}
\end{figure}

Uncertainty estimation and quantification are important tools for building trustworthy and reliable machine learning systems~\cite{lin2012survey,kabir2018neural,riedmaier2021unified}. 
Particularly, when such machine-learned systems are applied to make predictions that involve important decisions, e.g., medical diagnosis~\cite{ghoshal2020estimating}, financial planning and decision-making ~\cite{baker2020covid,oh2021managing}, and autonomous driving~\cite{hoel2020tactical}. 
The recent development of neural networks has shown excellent predictive performance in many domains. 
Among those, transformers, including the vanilla transformer~\citep{vaswani2017attention} and its variants such as BERT~\cite{devlin2019bert,wang2020position} are the representative state-of-the-art type of neural architectures that have shown remarkable performance on various recent Natural Language Processing (NLP)~\cite{gillioz2020overview} and Information Retrieval (IR)~\cite{ren2021conversations} tasks.  

Although transformers excel in terms of predictive performance~\cite{tetko2020state,han2021transformer},
they do not offer the opportunity for practitioners to inspect the model confidence due to their deterministic nature, i.e., incapability to assess if transformers are confident about their predictions. 
This influence is non-trivial because transformers are cutting-edge basic models for NLP.
Thus, estimating the predictive uncertainty of transformers benefits a lot in terms of building and examining model reliability for the downstream tasks. 

To estimate the uncertainty of neural models' prediction, one common way is to inject stochasticity (e.g., noise or randomness)~\cite{kabir2018neural,gawlikowski2021survey}.
It enables models to output a predictive distribution, instead of a single-point prediction. 
Casting a deterministic transformer to be stochastic requires us to take the training and inference computational complexity into consideration,
because uncertainty estimation usually relies on multiple forward runs. Therefore, directly adapting the aforementioned methods is not desired, given the huge amount of parameters and architectural complexity of transformers.

Figure~\ref{fig:uncertainty} outlines deterministic transformer (Figure~\ref{fig:uncertainty}(a)) and the possible approaches (Figure~\ref{fig:uncertainty}(b-e) for making a stochastic transformer.
\acs{BNN} (Figure~\ref{fig:uncertainty}(b)) assumes the network weights follow a Gaussian or a mixture of Gaussian~\citep{blundell2015weight}, and tries to learn the weight distribution $(\mu, \sigma)$, instead of weight $W$ itself, with the help of re-parameterization trick~\cite{kingma2013auto}. 
That means, BNN doubles the number of parameters. 
This is particularly challenging for a large network like a transformer, which has millions of parameters to be optimized.   
To alleviate this issue, \acs{MC} dropout~\cite{gal2016dropout} (Figure~\ref{fig:uncertainty}(c)) uses dropout~\cite{srivastava2014dropout}, concretely Bernoulli distributed random variables, to approximate the exact posterior distribution~\cite{gal2016dropout}.
However, \acs{MC} dropout tends to give overconfident uncertainty estimation~\cite{foong2019between}.
Ensemble~\cite{lakshminarayanan2017simple}(Figure~\ref{fig:uncertainty}(d)) is an alternative way to model uncertainty by averaging $N$ independently trained models, which yields the computational overhead by $N$ times in model training.

Unlike models above, we propose a simple yet effective approach based on Gumbel-Softmax tricks or Concrete Dropout~\cite{jang2017categorical,maddison2017the}, which are independently found for continuous relaxation, to estimate uncertainty of transformers. 
First, we cast the deterministic attention distribution for values in each self-attention head to be stochastic. 
The attention is then sampled from a Gumbel-Softmax distribution, which controls the concentration over values. 
Second, we regularize the key heads in self-attention to attend to a set of learnable centroids. 
This is equivalent to performing clustering over keys~\cite{vyas2020fast} or clustering hidden states in \acs{RNN}~\cite{wang2019state, wang2020uncertainty}. Similar attention mechanism has been also used to allow the layers in the encoder and decoder attend to inputs in the Set Transformer~\citep{lee2019set} and to estimate attentive matrices in the Capsule networks~\citep{ahmed2019star}.
Third, each new key head will be formed with a mixture of Gumbel-Softmax sampled centroids.
The stochasticity is injected by sampling from a Gumbel-Softmax distribution.
This is different from \acs{BNN} (sampling from Gaussian distribution), MC-dropout (sampling from Bernoulli distribution), Ensemble (the stochasticity comes from random seeds in model training).
With this proposed mechanism, we approximate the vanilla transformer with a stochastic transformer based on a hierarchical stochastic self-attention, namely $\textsc{H-sto-Trans}$, which enables the sampling of attention distributions over values as well as over a set of learnable centroids. 

Our work makes the following contributions: 
\begin{itemize}
\item We propose a novel way to cast the self-attention in transformers to be stochastic, which enables transformer models to provide uncertainty information with predictions. 
\item We theoretically show that the proposed self-attention approximation is upper bounded, the key attention heads that are close in Euclidean distance have similar attention distribution over centroids.   
\item In two benchmark tasks for \acs{NLP}, we empirically demonstrate that \textsc{H-sto-Trans} 
(1) achieves very competitive (in most cases, better) predictive performance on in-domain datasets; 
(2) is on par with baselines in uncertainty estimation on out-of-domain datasets;
(3) learns a better predictive performance-uncertainty trade-off than compared baselines, i.e., high predictive performance and low uncertainty on in-domain datasets, high predictive performance and high uncertainty on out-of-domain datasets. 
\end{itemize}
\begin{figure}
\label{fig:self-att}
\centering
\includegraphics[width=\columnwidth]{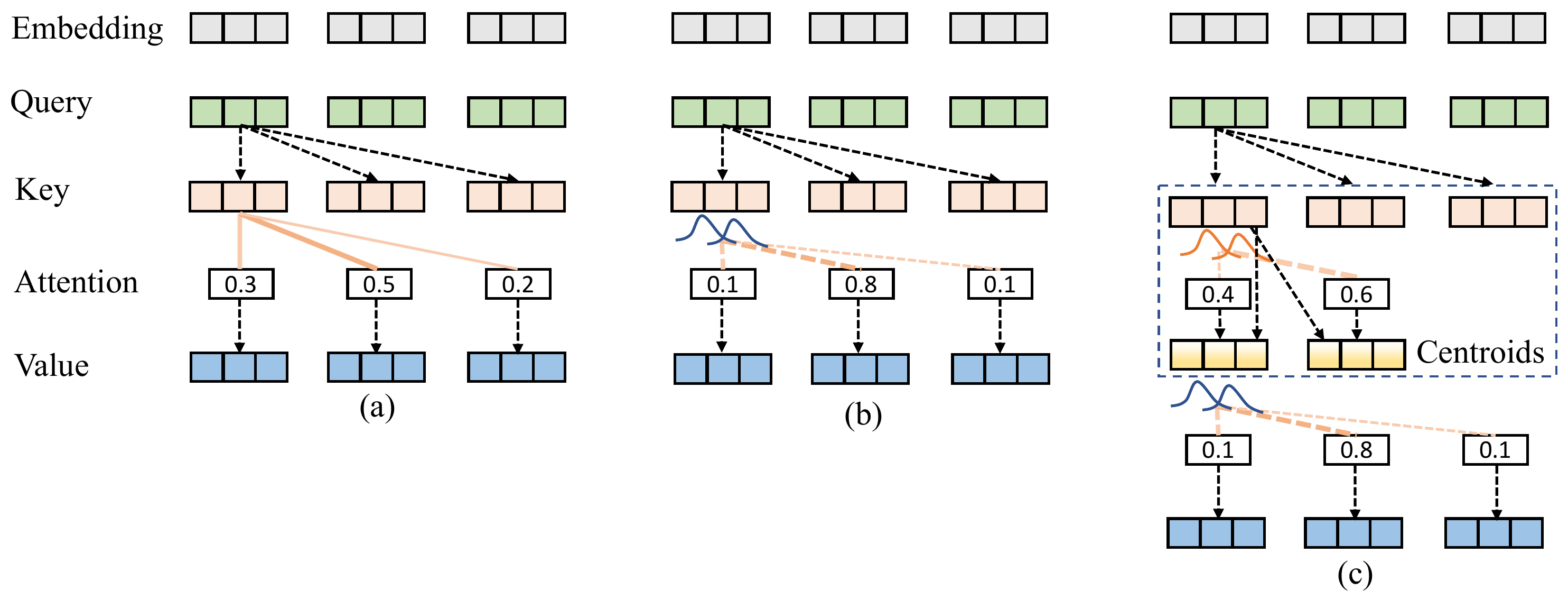}
\caption{The illustration of multi-head self-attention in deterministic and stochastic transformers. (a) The vanilla transformer with deterministic self-attention. (b) Stochastic transformer has stochastic self-attention used to weight values $V$, the standard Softmax is replaced with the Gumbel-Softmax. (c) Hierarchical stochastic transformer learns to pay attention to values $V$ and a set of learnable centroids $C$ stochastically.}
\vspace*{-1\baselineskip}
\end{figure}

\section{BACKGROUND}
\subsection{Predictive Uncertainty}
The predictive uncertainty estimation is a challenging and unsolved problem. 
It has many faces, depending on different classification rules. 
Commonly it is classified as epistemic (model) and aleatoric (data) uncertainty~\citep{der2009aleatory,kendall2017uncertainties}. 
Alternatively, on the basis of input data domain, it can also be classified into \textit{in-domain (ID)}~\citep{ashukha2019pitfalls} and \textit{out-of-domain (OOD)} uncertainty~\citep{hendrycks2016baseline,wang2020doubly}. 
With in-domain data, i.e. the input data distribution is similar to training data distribution, a reliable model should exhibit high predictive performance (e.g., high accuracy or F1-score) and report high confidence (low uncertainty) on correct predictions. 
On the contrary, out-of-domain data has quite different distribution from training data, an ideal model should give high predictive performance to illustrate the generalization to unseen data distribution, but desired to be unconfident (high uncertainty). 
We discuss the epistemic (model) uncertainty in the context of ID and OOD scenarios in this work. 

\subsection{Vanilla Transformer}
\label{sec:trans}
The vanilla transformer~\cite{vaswani2017attention} is an alternative architecture to Recurrent Neural Networks \acfp{RNN} for modelling sequential data that relaxes the model’s reliance on input sequence order. 
It consists of multiple components such as positional embedding, residual connection and multi-head scaled dot-product attention. 
The core component of the transformer is the multi-head self-attention mechanism. 

Let $\bfx \in \bR^{l \times d}$ ($l$ is sequence length, $d$ is dimension) be input data, and $\bW_q, \bW_k, \bW_v \in \bR^{d \times d}$ be the matrices for query $Q \in \bR^{l\times h \times d_h}$, key $K \in \bR^{l\times h \times d_h}$, and value $V \in \bR^{l\times h \times d_h}$, $d_h=\frac{d}{h}$ and $h$ is the number of attention heads. 
Each $\bfx$ is associated with a query $Q$ and a key-value pair $(K, V)$. 
The computation of an attentive representation $A$ of $\bfx$ in the multi-head self-attention is:
\begin{align}
Q&=\bW_q \bfx; ~~~~~~
K=\bW_k \bfx; ~~~~~~
V=\bW_v \bfx; ~~~~ \\
A&=\textsc{Softmax}(\alpha^{-1}QK^\top);~~~~H =A V 
\label{eq:sto_trans}
\end{align}
where $H =[h_1,...,h_h]$ is the multi-head output and $A=[a_1,...,a_h]$ is the attention distribution that needs to attend to $V$, $\alpha$ is a scaling factor.
Note that a large value of $\alpha$ pushes the Softmax function into regions where it has extremely small gradients.
This attention mechanism is the key factor of a transformer for achieving high computational efficiency and excellent predictive performance. 
However, as we can see, all computation paths in this self-attention mechanism are deterministic, leading to a single-point output. 
This limits us to access and evaluate the uncertainty information beyond prediction given an input $\bfx$. 

We argue that the examination of the reliability and confidence of a transformer prediction is crucial for many \acs{NLP} applications, particularly when the output of a model is directly used to serve customer requests.
In the following section, we introduce a simple yet efficient way to cast the deterministic attention to be stochastic for uncertainty estimation based on Gumbel-Softmax tricks~\cite{jang2017categorical,maddison2017the}. 

\section{METHODOLOGY}
\subsection{Bayesian Inference and Uncertainty Modeling}
In this work, we focus on using transformers in classification tasks. 
Let $D=\{X, Y\}=\{x_i, y_i\}_{i=1}^N$ be a training dataset, $y_i \in \{1,...,M\}$ is the categorical label for an input $x_i \in \bR^d$. 
The goal is to learn a transformation function $f$, which is parameterized by weights $\omega$ and maps a given input $x$ to a categorical distribution $y$. 
The learning objective is to minimize negative log likelihood, $\mathcal{L} = -\frac{1}{N}\sum_i^N \log p(y_i|x_i, \omega)$. 
The probability distribution is obtained by Softmax function as:
\begin{align}
p(y_i=m|x_i, \omega)=\frac{\exp(f_m(x_i, \omega))}{\sum_{k \in M}\exp(f_k(x_i, \omega)}. 
\end{align}
In the inference phase, given a test sample $x^*$, the predictive probability $y^*$ is computed by:
 \begin{align}
p(y^*|x^*, D)= \int p(y^*|x^*, \omega) p(\omega|D)d\omega
\end{align}
where the posterior $p(\omega|D)$ is intractable and cannot be computed analytically. 
A variational posterior distribution $q_\theta(\omega)$, where $\theta$ are the variational parameters, is used to approximate the true posterior distribution by minimizing the Kullback-Leilber (KL) distance. 
This can also be treated as the maximization of \acf{ELBO}:
 \begin{align}
\mathcal{L_\theta}=\int q_\theta(\omega)p(Y|X, \omega)d\omega-\textbf{KL}[q_\theta(\omega)\parallel p(\omega)]
\end{align}
With the re-parametrization trick~\cite{kingma2015variational}, a differentiable mini-batched Monte Carlo estimator can be obtained. 

The predictive (epistemic) uncertainty can be measured by performing $T$ inference runs and averaging predictions. 
\begin{align}
p(y*|x*)= \frac{1}{T}\sum_{t=1}^T p_{\omega_t}(y^*|x^*, \omega_t)
\end{align}

$T$ corresponds to the number of sets of mask vectors from Bernoulli distribution $\{r^t\}_{t=1}^T$ in MC-dropout, or the number of randomly trained models in Ensemble, which potentially leads to different set of learned parameters $\omega=\{\omega_1,...,\omega_t\}$, or the number of sets of sampled attention distribution from Gumbel distribution $\{g^t\}_{t=1}^T$ in our proposed method.

\subsection{Stochastic Self-Attention with Gumbel-Softmax} 
As described in section \ref{sec:trans}, the core component that makes a transformer successful is the multi-head self-attention. 
For each $i$-th head, let $q_i \in Q, k_i \in K,  v_i \in V$, it is written as: 
\begin{align}
a_i &= \textsc{Softmax}(\frac{q_i k_i^\top}{\tau }); ~~a_i \in \bR^{l\times l} \\
h_i &= a_i v_i; ~~~~h_i \in \bR^{l\times d_h}
\end{align} 
We here use a temperature parameter $\tau$ to replace the scaling factor $\alpha$. The $a_i$ is attention distribution, which learns the compatibility scores between tokens in the sequence with the $i$-th attention head. The scores are used to retrieve and form the mixture of the content of values, which is a kind of content-based addressing mechanism in neural Turing machine~\citep{graves2014neural}. Note the attention is deterministic.

A straightforward way to inject stochasticity is to replace standard Softmax with Gumbel-Softmax, which helps to sample attention weights to form $\hat{a}_i$.
\begin{align}
\label{eq:s_attn_a}
\hat{a_i} &\sim  \mathcal{G}(\frac{q_i k_i^\top}{\tau })\\
\label{eq:s_attn_h}
h_i &= \hat{a_i} v_i
\end{align}
where $\mathcal{G}$ is \textsc{Gumbel-Softmax} function. The Gumbel-Softmax trick is an instance of a path-wise Monte-Carlo gradient estimator~\cite{gumbel1954statistical,maddison2017the,jang2017categorical}. 
With the Gumbel trick, we can draw samples $z$ from a categorical distribution given by parameters $\bm{\theta}$, that is, 
$\bm{z}= \mbox{\textsc{one\_hot}} \big( \argmax_i [g_i + \log\theta_i] \big), i \in [1\dots k]$,
where $k$ is the number of categories and $g_i$ are i.i.d. samples from the \textsc{Gumbel}$(0, 1)$, that is, $g=-\log(-\log(u)), u \sim \textsc{uniform}(0, 1)$ is independent to network parameters. 
Because the $\argmax$ operator breaks end-to-end differentiability, the categorical distribution $\bm{z}$ can be approximated using the differentiable Softmax function \cite{jang2017categorical,maddison2017the}. Here the $\tau$ is a tunable temperature parameter equivalent to $\alpha$ in Eq. (\ref{eq:sto_trans}), Then the attention weights (scores) for values  in Eq.\ref{eq:sto_trans} can be computed as:
\begin{equation}
\hat{a}_i= \frac{\exp ((\log(\theta_{i}) +g_i)/\tau)}{\sum_{j=1}^{k} \exp ((\log(\theta_{t_j})+g_j)/\tau)}, ~~i \in [1\dots k].
\label{eq:diffrentiable_approxmiation}
\end{equation} 
where the $\theta_{i}=q_ik_i^\top$. 
And we use the following approximation:
\begin{equation}
\textbf{KL}[a \parallel  \hat{a}]~~~\text{where} ~~a_j = \frac{a_j}{\sum_{k}^{i=1}a_i}
\end{equation} 

This indicates an approximation of a deterministic attention distribution $a$ with a stochastic attention distribution $\hat{a}$. With a larger $\tau$, the distribution of attention is more uniform, and with a smaller $\tau$, the attention becomes more sparse. 

\subsubsection{The trade-off between predictive performance and uncertainty estimation.} 
This trade-off is rooted in bias-variance trade-off. 
Let $\phi (x)$ be a prediction function, and $f(x)$ is the true function and $\rho$ be a constant number. 
The error can be computed as:
\begin{equation}
\xi(x) = \underbrace{(\mathbb{E}[\phi (x)-f(x)])^2}_{Bias^2} + \underbrace{(\mathbb{E}[\phi (x)-E[\phi (x)]]^2)}_{Variance} + \underbrace{\rho}_{Const}
\end{equation} 

\acs{MC}-dropout~\cite{gal2016dropout} with $T$ times \acl{MC} estimation gives a prediction $\mathbb{E}[\phi_t (x)], t \in T$ and predictive uncertainty, e.g., variances $Variance[\phi_t (x)]$ ($\rho$ is a constant number denotes irreducible error). 
On both in-domain and out-of-domain datasets, a good model should exhibit low bias, which ensures model generalization capability and high predictive performance. For epistemic (model) uncertainty, we expect model outputs low variance on in-domain data and high variance on out-of-domain data. 

We empirically observe (from Table \ref{tab:imdb_results} and Table \ref{tabLcola}) that this simple modification in Eq. (\ref{eq:s_attn_a}) can effectively capture the model uncertainty, but it struggles to learn a good trade-off between predictive performance and uncertainty estimation. 
That is, when good uncertainty estimation performance is achieved on out-of-domain data, the predictive performance on in-domain data degrades. 
To address this issue, we propose a hierarchical stochastic self-attention mechanism.

\subsection{Hierarchical Stochastic Self-Attention}
To further encourage transformer model to have stochasticity and retain predictive performance, we propose to add an additional stochastic attention before the attention that pays values.
This attention forces each key head stochastically attend to a set of learnable centroids, which will be learned during back-propagation. 
This is equivalent to regularizing key attention heads. 
Similar ideas have been used to improve transformer efficiency~\citep{vyas2020fast} and to improve \acs{RNN} memorization~\citep{wang2019state}. 

We first define the set of $c$ centroids, $C \in \bR^{d_h \times c}$ . 
Let each centroid $c_i \in \bR^ {d_h}$ have the same dimension with each key head $k_j \in \bR^{d_h}$. 
The model will first learn to pay attention to centroids, and a new key head $\hat{k}_j$ is formed by weighting each centroid. 
Then $\hat{k}$ and a query $q$ decides the attention weights to combine values $v$. 
For the $i$-th head, a given query $q_i$, key $k_i$, value $v_i$, the stochastic self-attention can be hierarchically formulated as:
\begin{align}
\label{eq:hs_attn_ac}
\hat{a}_{c} &\sim \mathcal{G}(\tau_1^{-1}k_iC), ~~\hat{a}_{c} \in \bR^{l  \times c}\\
\label{eq:hs_attn_k}
\hat{k_i} &= \hat{a}_c C^\top, ~~\hat{k_i} \in \bR^{l \times d_h} \\
\label{eq:hs_attn_av}
\hat{a}_v &\sim \mathcal{G}(\tau_2^{-1}q_i\hat{k}_i^\top), ~~\hat{a}_v \in \bR^{l \times l}  \\
\label{eq:hs_attn_h}
h_i &= \hat{a}_v v_i
\end{align}
$\hat{a}_{c}, \hat{a}_{v}$  are the sampled categorical distributions that are used to weight centroids in $C$ and tokens in $v_i$. The $\tau_1, \tau_2$ control the softness for each stochastic self-attention, respectively. 

We summarize the main procedures of performing hierarchical stochastic attention in transformer in Algorithm~\ref{alg:hs-attn}. 
\begin{algorithm}
    \small
    \SetAlgoLined
    \SetKwInOut{Input}{\textbf{Input}}\SetKwInOut{Output}{\textbf{Output}} 
    \Input{query $Q$,  key $K$, value $V$, centroids $C$}
    \Output{Hierarchical stochastic attentive output $H$}
    \BlankLine
    Model stochastic attention $\hat{A_c}$ over centroids $C$ as Eq.\ref{eq:hs_attn_ac};\\
    Sample $\hat{A}_c$ from a categorical distribution $\bm{z}= \mbox{\textsc{one\_hot}} \big( \argmax_i [g_i + \log\theta_i] \big), i \in [1\dots k]$, $g=-\log(-\log(u)), u \sim \textsc{uniform}(0, 1)$ ;\\
    Differentially approximate $\hat{A}_c$ as Eq.~\ref{eq:diffrentiable_approxmiation};\\
    Compute $\hat{K}=\hat{A}_cC^{\top}$ as Eq.~\ref{eq:hs_attn_k};\\
    Model stochastic attention $\hat{A_v}$ over value $V$ as Eq.\ref{eq:hs_attn_av};\\
    Sample and approximate $\hat{A_v}$, similar to line 2 to 3;\\ 
    Compute $H=\hat{A}_vV$ as Eq.~\ref{eq:hs_attn_h};\\
    \caption{Hierarchical stochastic transformer.}
    \label{alg:hs-attn}
\end{algorithm}

\subsubsection{Why perform clustering on key heads?}
The equation (\ref{eq:hs_attn_ac}) performs clustering on the key attention heads and outputs an attention distribution, and equation (\ref{eq:hs_attn_k}) tries to form a new head based on attention distribution and learned centroids. 
The goal is to make the original key heads to be stochastic, allowing attention distribution to have randomness for uncertainty estimation. 
This goal can be also accompanied by applying equations (\ref{eq:hs_attn_ac}) and (\ref{eq:hs_attn_k}) to query while keeping key unchanged.  In that case, $\hat{a}_{c}$ can be still sampled stochastically based on query and centroids.

\subsubsection{Stochastic attention approximation.}
The equations (\ref{eq:hs_attn_ac}) and (\ref{eq:hs_attn_k}) group the key heads into a fixed number of centroids and are reweighed by the mixture of centroids. 
As in~\citep{vyas2020fast}, we can analyze the attention approximation error, and derive that the key head attention difference is bounded.
\begin{proposition}
\label{prop:att}
Given two keys $k_i$ and $k_j$ such that $\left \| k_i-k_j \right \|_2 \leq\varepsilon$, stochastic key attention difference is bounded: $\left \|\mathcal{G}(\tau^{-1}k_i C))-\mathcal{G}(\tau^{-1}k_j C)) \right \|_2 \leq\tau^{-1}\varepsilon \left \| C \right \|_2$, where $\mathcal{G}$ is the Gumbel-Softmax function, and $\left \| C \right \|_2$ is the spectral norm of centroids. 
$\varepsilon$ and $\tau$ are constant numbers.
\end{proposition}

\begin{proof}
Same to the Softmax function, which has Lipschitz constant less than 1~\citep{gao2017properties}, we have the following derivation:
\begin{equation}
 \begin{aligned}
 &\left \|\mathcal{G}(\tau^{-1}k_i C))-\mathcal{G}(\tau^{-1}k_j C)) \right \|_2\\
 &~\leq \left \|\tau^{-1}k_i C-\tau^{-1}k_j C \right \|_2\\
 &~\leq \tau^{-1}\varepsilon \left \| C \right \|_2
\end{aligned}   
\end{equation}
\end{proof}

Proposition~\ref{prop:att} shows that the $i$-th key assigned to $j$-th centroid can be bounded by its distance from $j$-th centroid. 
The keys that are close in Euclidean space have similar attention distribution over centroids. 

\section{EXPERIMENTAL SETUPS}
We design experiments to achieve the following objectives:
\begin{itemize}
\item To evaluate the predictive performance of models on in-domain datasets. High predictive scores and low uncertainty scores are desired.
\item To compare the model generalization from in-domain to out-of-domain datasets. High scores are desired.
\item To estimate the uncertainty of the models on out-of-domain datasets. High uncertainty scores are desired.
\item To measure the model capability in learning the predictive performance and uncertainty estimation trade-off. 
\end{itemize}

\subsection{Datasets}
We use IMDB dataset\footnote{https://ai.stanford.edu/~amaas/data/sentiment/}~\cite{maas2011learning} for the sentiment analysis task. 
The standard IMDB has 25,000/25,000 reviews for training and test, covering 72,062 unique words. For hyperparameter selection, we take 10\% of training data as validation set, leading to 22,500/2,500/25,000 data samples for training, validation, and testing. 
Besides, we use \acf{CR} dataset~\cite{hendrycks2016baseline} which has 500 samples to evaluate the proposed model in OOD settings.
We conduct the second experiment on linguistic acceptability task with CoLA dataset\footnote{https://nyu-mll.github.io/CoLA/}~\cite{warstadt2019neural}. 
It consists of 8,551 training and 527 validation in-domain samples. As the labels of test set is not publicly available,
we split randomly the 9078 in-domain samples into train/valid/test with 7:1:2. 
Additionally, we use the provided 516 out-of-domain samples for uncertainty estimation. 

\subsection{Compared Methods.} 
We compare the following methods in our experimental setup: 
\begin{itemize}
\item \textsc{Trans}~\cite{vaswani2017attention}: The vanilla transformer with deterministic self-attention.
\item \textsc{MC-dropout}~\cite{gal2016dropout}: Using dropout~\citep{srivastava2014dropout} as a regularizer to measure the prediction uncertainty.
\item \textsc{Ensemble}~\cite{lakshminarayanan2017simple}: Average over multiple independently trained transformers.
\item \textsc{sto-trans}: The proposed method that the attention distribution over values is stochastic; 
\item \textsc{H-sto-tran}: The proposed method that uses hierarchical stochastic self-attention, i.e., the stochastic attention from key heads to a learnable set of centroids and the stochastic attention to value, respectively.
\end{itemize}

\subsection{Implementation details}
We implement models in PyTorch~\cite{paszke2019pytorch}. 
The models are trained with Adam~\cite{kingma2014adam} as the optimization algorithm. 
For each trained model, we sample 10 predictions (run inference 10 times), the mean and variance (or standard deviation) of results are reported. The uncertainty information is quantified with variance (or standard deviation).
For sentiment analysis, we use 1 layer with 8 heads, both the embedding size and the hidden dimension size are 128. 
We train the model with learning rate of 1e-3, batch size of 128, and dropout rate of 0.5/0.1. 
We evaluate models at each epoch, and the models are trained with maximum 50 epochs.
We report accuracy as the evaluation metric. 
For linguistic acceptability, we use 8 layers and 8 heads, the embedding size is 128 and the hidden dimension is 512. 
We train the model with learning rate of 5e-5, batch size of 32 and dropout rate of 0.1. 
We train the models with maximum 2000 epochs and evaluate the models at every 50 epochs.
We use \acf{MCC}~\cite{matthews1975comparison} as the evaluation metric. 
The model selection is performed based on validation dataset according to predictive performance.

\section{EXPERIMENTAL RESULTS}
\subsection{Results on Sentiment Analysis}

\begin{figure*}[!htb]
\centering
\includegraphics[width=0.66\textwidth]{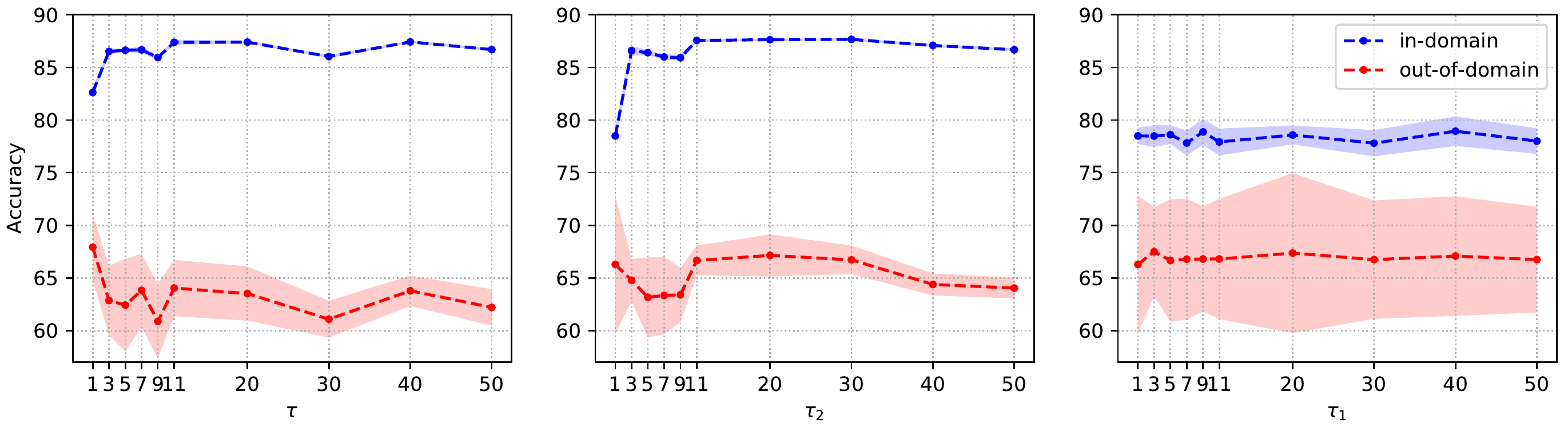}
\caption{The experiments with hyperparameter $\tau$. Left: \textsc{sto-trans} with different $\tau$. The randomness is solely based on the sampling on attention distribution over values. While uncertainty information is captured, \textsc{sto-trans} has difficulties in learning the trade-off between in-domain and out-of-domain performance. Middle: The hyperparameter tuning of $\tau_1$ and $\tau_2$ in \textsc{h-sto-trans}. $\tau_1$ controls the concentration on centroids and $\tau_2$ controls the concentration on values.}
\label{fig:self-att-exp}

\end{figure*}

\begin{table*}[!htb]
\centering
\footnotesize
\begin{tabular}{@{}lcccc@{}}
\hline
             & ID (\%) & OOD (\%) & $\bigtriangledown$ID (\%)  &$\bigtriangledown$OOD (\%) \\ \hline \hline

\textsc{trans} ($\eta=0.1$) & 87.00  & 65.00 & / & /\\ 
\textsc{trans} ($\eta=0.5$)  & 87.51  & 63.40 & 0.51 $\uparrow$& 1.60 $\downarrow$\\ 
\textsc{MC-dropout} ($\eta=0.5$) & 86.06 $\pm$ 0.087 & 63.38 $\pm$ 1.738 &0.94 $\uparrow$& 1.62 $\downarrow$ \\  
\textsc{MC-dropout} ($\eta=0.1$) & 87.01 $\pm$ 0.075 & 63.38 $\pm$ 0.761 &0.10 $\uparrow$& 1.62 $\downarrow$ \\ 
\textsc{ensemble} & 86.89 $\pm$ 0.230& 64.20 $\pm$ 1.585 &0.11 $\downarrow$& 0.80 $\downarrow$\\ \hline
\textsc{sto-trans} ($\tau=1$) & 82.62 $\pm$ 0.092 & 67.92 $\pm$ 0.634 &4.38 $\downarrow$& 2.92 $\uparrow$ \\ %
\textsc{sto-trans} ($\tau=40$) & 87.42 $\pm$ 0.022 & 63.78 $\pm$ 0.289 &0.42 $\uparrow$& 1.22 $\downarrow$\\ %
\textsc{h-sto-trans}  ($\tau_1=1$, $\tau_2=20$) & 87.63 $\pm$ 0.017 & 67.14 $\pm$ 0.400 &0.63 $\uparrow$& 2.14 $\uparrow$ \\ %
\textsc{h-sto-trans}  ($\tau_1=1$, $\tau_2=30$) & 87.66 $\pm$ 0.022 & 66.72 $\pm$ 0.271 &0.66 $\uparrow$& 1.72 $\uparrow$\\ %
\hline
\end{tabular}
\caption{The predictive performance and uncertainty estimation of models on IMDB (ID) and CR (OOD) dataset. The uncertainty estimation is performed by running forward pass inference by 10 runs, then the uncertainty is quantified by standard deviation across runs. For ensemble, the results are averaged on 10 models that are independently trained with random seeds. Dropout is used in the inference of \textsc{MC-dropout} and $\eta$ is dropout rate. In the rest of methods, dropout is not used in inference. The $\bigtriangledown$ID (\%) and $\bigtriangledown$OOD (\%) present the predictive performance difference to \textsc{trans} ($\eta=0.1$).  }
\label{tab:imdb_results}
\vspace*{-1\baselineskip}
\end{table*}
Table \ref{tab:imdb_results} presents the predictive performance and uncertainty estimation on IMDB (in-domain, ID) and CR (out-of-domain, OOD) dataset, evaluated by accuracy. 

First, \textsc{sto-trans} and \textsc{h-sto-trans} are able to provide uncertainty information, as well as maintain and even slightly outperform the predictive performance of \textsc{trans}. 
Specially, \textsc{sto-trans} ($\tau=40$) and \textsc{h-sto-trans} ($\tau_1=1$, $\tau_2=30$) outperforms \textsc{trans} ($\eta=0.1$) by 0.42\% and 0.66\% on ID dataset.
In addition, they allow us to measure the uncertainty via predictive variances.
It is because they inject randomness directly to self-attentions.
However, \textsc{trans} has no access to uncertainty information due to its deterministic nature.

Second, \textsc{sto-trans} is struggling to learn a good trade-off between ID predictive performance and OOD uncertainty estimation performance.
With small temperature $\tau=1$, \textsc{sto-trans} gives good uncertainty information, but we observe that the ID predictive performance drops. 
When $\tau$ approaches to $\sqrt{d/h}$ (the original scaling factor in the vanilla transformer), \textsc{sto-trans} achieves better performance on ID dataset, but lower performance on OOD dataset.
We conjecture that the randomness in \textsc{sto-trans} is solely based on the attention distribution over values and is not enough for learning the trade-off. 

Third, \textsc{h-sto-trans} achieves better accuracy-uncertainty trade-off compared with \textsc{sto-trans}. 
For instance, with $\tau_1=1, \tau_2=20$, \textsc{h-sto-trans} achieves 87.63\% and 67.14\%, which outperform the corresponding numbers of \textsc{sto-trans} for both ID and OOD datasets. It also outperforms \textsc{MC-dropout} and \textsc{ensemble},
specially, \textsc{h-sto-trans} outperforms 0.62\%-1.6\% and 2.52\%-3.76\% on ID and OOD datasets, respectively.
On OOD dataset, while \textsc{MC-dropout} and \textsc{ensemble} exhibit higher uncertainty (measured by standard deviation) across runs, the accuracy is lower than that of  \textsc{trans} ($\eta=0.1$), \textsc{sto-trans} ($\tau=1$) and \textsc{h-sto-trans}. 
It is due to a better way of learning two types of randomness: one from sampling over a set of learnable centroids and the other one from sampling attention over values. 

Figure \ref{fig:self-att-exp} reports the hyperparameter tuning of $\tau_1$ and $\tau_2$. 
The goal is to find a reasonable combination to achieve high predictive performance on both ID and OOD datasets. 
To simplify the tuning work, we fix the $\tau_1=1$ and then change $\tau_2$ with different values, and vice versa. 
As we can see, the combination of a small $\tau_1$ and a large $\tau_2$ performs better than the other way around. 
We think this is because $\tau_2$ is in the latter stage and has bigger effects on the predictive performance. 
However, removing $\tau_1$ goes back to Figure \ref{fig:self-att-exp} (Left), where accuracy-uncertainty trade-off is not well learned by \textsc{sto-trans}.

\subsection{Results on Linguistic Acceptability}
Table \ref{tabLcola} shows the performance of compared models on both in-domain (ID) and out-of-domain (OOD) sets of CoLA dataset, evaluated by \acs{MCC}.
\begin{table}[!htb]
    \centering
    \scriptsize
    \setlength{\tabcolsep}{2pt}
    \begin{tabular}{lcccc} \hline
     Models & ID(\%) & OOD(\%) & $\bigtriangledown$ID (\%)  &$\bigtriangledown$OOD (\%) \\ \hline \hline
    \textsc{Trans} ($\eta=0.1$) & 20.09 & 16.46 & /& /\\
    \textsc{MC-dropout}  ($\eta=0.1$) & 19.91 $\pm$  0.40 & 16.70 $\pm$ 2.21 & 0.18 $\downarrow$&0.24 $\uparrow$ \\
    \textsc{MC-dropout} ($\eta=0.05$) & 20.03 $\pm$  0.30 & 17.11 $\pm$ 1.21 & 0.06 $\downarrow$&0.65 $\uparrow$\\
    \textsc{Ensemble} & 21.20 $\pm$  2.59 & 16.73 $\pm$ 4.92 & 1.11 $\uparrow$&0.27  $\uparrow$\\ \hline
    \textsc{sto-trans} & 23.27 $\pm$  0.75 & 15.25 $\pm$ 4.65 & 3.18 $\uparrow$&1.21 $\downarrow$\\
    \textsc{h-sto-trans} & 20.52 $\pm$  0.76 & 16.49 $\pm$ 4.08 & 0.43 $\uparrow$&0.03  $\uparrow$\\ \hline
    \end{tabular}
    \caption{The performance of compared models on CoLA dataset. We set all temperature values $\tau_1=1$ and $\tau_2=1$. The $\bigtriangledown$ID (\%) and $\bigtriangledown$OOD (\%) present the predictive performance and difference to \textsc{trans} ($\eta=0.1$), respectively.}
    \label{tabLcola}
\end{table}

\begin{table}[!htb]
    \scriptsize
    \center
    \setlength{\tabcolsep}{0.1pt}
    \begin{tabular}{llr} \\ \hline
    Examples (Labels) & Prob. Corr. & Corr./Total  \\ \hline \hline
    no man has ever beaten the centaur.  (1) & 0.75 $\pm$ 0.001 & 10/10   \\
    nora sent the book to london (1) & 0.65 $\pm$ 0.007 & 10/10    \\
    sally suspected joe, but he did n't holly. (1)& 0.60 $\pm$ 0.008 & 8/10    \\
    kim is eager to recommend. (0) & 0.41 $\pm$ 0.011 & 3/10  \\
    he analysis her was flawed (0)& 0.24 $\pm$ 0.003 & 0/10     \\ \hline 
    sandy had read how many papers ? ! (1) & 0.67 $\pm$ 0.010 & 10/10   \\
    which book did each author recommend ? (1) & 0.58 $\pm$ 0.010 & 7/10    \\
    she talked to harry , but i do n't know who else .(1)  & 0.52 $\pm$ 0.013 & 4/10   \\
    john is tall on several occasions . (0)& 0.42 $\pm$ 0.005 & 1/10   \\
    they noticed the painting , but i do n't know for how long . (0)& 0.28 $\pm$ 0.003 & 0/10   \\ \hline
    \end{tabular}
    \caption{Illustration of predictions with \textsc{h-sto-trans}. The predictions for each ID (top) and OOD (bottom) samples are measured by the probability of being correct of each prediction and the number of correct predictions. }
    \label{tab:example}
\end{table}

First, \textsc{sto-trans} and \textsc{h-sto-trans} obtain comparable performance as well as provide uncertainty information, compared with \textsc{trans}.
To be specific, \textsc{sto-trans} and \textsc{h-sto-trans} improves 3.18\%  and 0.43\% of \acs{MCC} on ID dataset compared with deterministic \textsc{trans} respectively.

Second, \textsc{sto-trans} achieves the best performance on ID dataset but the worst performance on OOD dataset.
Although \textsc{sto-trans} outperforms \textsc{trans}, the best \textsc{MC-dropout}, \textsc{ensemble} by 3.18\%, 3.24\%, 2.07\%  of MCC on ID dataset,
its performance drops by 1.21\%, 1.86\%, 1.48\%, correspondingly on OOD dataset. 
This further verifies our conjecture that the randomness is only introduced to attention distribution over values and is insufficient for learning the trade-off of ID and OOD data.

Third, \textsc{h-sto-trans} enabled to learn better trade-off between prediction and uncertainty.
Precisely, the performance improves 0.43\% and 0.03\% of MCC on ID and OOD datasets respectively.
\textsc{h-sto-trans} is 0.49\% superior to \textsc{MC-dropout} ($\eta=0.05$), meanwhile, 0.68\% inferior to \textsc{ensemble} on ID dataset.
Given \textsc{ensemble} shows high uncertainty on ID dataset and \textsc{MC-dropout} ($\eta=0.05$) has low uncertainty on OOD dataset, this is not desired. 
Therefore, \textsc{h-sto-trans} is the one that strikes the better balance across the objectives. In the context of this task, it means high MCC, low variance on ID dataset and high MCC, high variance on OOD dataset.

Table \ref{tab:example} gives some predictions of test samples with \textsc{h-sto-trans}. 
What we observed are two folds: 
(1) In general, ID predictions have lower variances in terms of the probability of being correct.
For ``10/10'' (10 correct predictions out of 10 total predictions) prediction cases, the ID examples have higher probability score than the ones in OOD data. 
Also, we find there are much less number of ``10/10'' prediction cases in OOD dataset than that in ID dataset.
(2) For ID dataset, either with high or low probability scores, we can see low variances, we see more ``10/10" (tend to be confidently correct) or ``0/10'' (tend to be confidently incorrect) cases.
As expected, for both cases, the variance is relatively low as compared to probability around 0.5.  
In deterministic models, we are not able to access this kind of information which would imply how confident are the transformer models towards predictions.

\section{RELATED WORK}
Bayesian neural networks~\cite{blundell2015weight} inject stochasticity by sampling the network parameters from a Gaussian prior. 
Then the posterior distribution of target can be estimated in multiple sampling runs.  
However, the Bayesian approach doubles the number of network parameters, i.e., instead of learning a single-point network parameter, it learns a weight distribution which is assumed to follow a Gaussian distribution. 
Additionally, it often requires intensive tuning work on Gaussian mean and variance to achieve stable learning curves as well as predictive performance. 
\acs{MC} dropout~\cite{gal2016dropout} approximates Bayesian approach by sampling dropout masks from a Bernoulli distribution. 
However, \acs{MC} dropout has been demonstrated to give overconfident uncertainty estimation~\cite{foong2019between}. 
Alternatively, the recently proposed deep ensembles~\cite{lakshminarayanan2017simple} offers possibility to estimate predictive uncertainty by combining predictions from different models which are trained with different random seeds. 
This, however, significantly increases the computational overhead for training and inference. 
There are some \acs{MC} dropout based methods recently proposed. 
Sequential \acs{MC} transformer~\citep{martin2020monte}, which models uncertainty by casting self-attention parameters as unobserved latent states by evolving randomly through time. ~\cite{he2020towards} combined mix-up, self-ensembling and dropout to achieve more accurate uncertainty score for text classification.
~\cite{shelmanov2021certain} proposed to incorporate determinantal point process (DPP) to \acs{MC} dropout to quantify the uncertainty of transformers. 
Different to the above-mentioned approaches, we inject stochasticity into the vanilla transformer with Gumbel-Softmax tricks. 
As it is shown in the experiment section, hierarchical stochastic self-attention component can effectively capture model uncertainty, and learn a good trade-off between in-domain predictive performance and out-of-domain uncertainty estimation.

\section{DISCUSSION}

While many extension of transformers have been recently proposed, the most of transformer variants are still deterministic. Our goal in this work is to equip transformers in a stochastic way to estimate uncertainty while retaining the original predictive performance. 
This requires special design in order to achieve the two goals without adding a major computational overhead to model training and inference like Ensembles and Bayesian Neural Network (BNN). The complexity gain of our method to its deterministic version is modest and requires an additional matrix $C \in \mathbb{R}^{d_h \times c}$. This is more efficient than Ensemble and BNN, which gives N ($N\geq 2$ for Ensemble and  $N= 2$ for BNN) times more weights.

\section{CONCLUSION}

This work proposes a novel, simple yet effective way to enable transformers with uncertainty estimation, as an alternative to \acs{MC} dropout and ensembles. 
We propose variants of transformers based on two stochastic self-attention mechanisms: 
(1) injecting stochasticity into the stochastic attention over values;
(2) forcing key heads to pay stochastic attention to a set of learnable centroids.
Our experimental results show that the proposed approach learns good trade-offs between in-domain predictive performance and out-of-domain uncertainty estimation performance on two \acs{NLP} benchmark tasks, and outperforms baselines.

\newpage
\bibliography{aaai22}

\end{document}